\documentclass[lettersize,journal]{IEEEtran}
\usepackage{amsmath,amsfonts}
\usepackage{algorithmic}
\usepackage{algorithm}
\usepackage{array}
\usepackage[caption=false,font=normalsize,labelfont=sf,textfont=sf]{subfig}
\usepackage{textcomp}
\usepackage{stfloats}
\usepackage{url}
\usepackage{verbatim}
\usepackage{graphicx}
\usepackage{cite}
\usepackage{xcolor}
\usepackage{amsmath}
\usepackage{amssymb}
\usepackage{booktabs}
\usepackage{multirow}
\usepackage{makecell}
\usepackage{color}
\usepackage{threeparttable}
\usepackage{orcidlink}
\hypersetup{hidelinks}
\usepackage{soul}
\soulregister\ref7 
\soulregister\cite7 

\newcommand{\orcidauthorA}{0009-0009-4073-8257}
\newcommand{\orcidauthorB}{0000-0001-9429-1096}
\newcommand{\orcidauthorC}{0000-0003-2847-0349}
\newcommand{\orcidauthorD}{0000-0002-7976-8439}
\newcommand{\orcidauthorE}{0000-0002-8581-9554}

\begin{document}

\title{StructDiff: A Structure-Preserving and Spatially Controllable\\ Diffusion Model for Single-Image Generation}
\author{
    Yinxi He$^{\orcidlink{\orcidauthorA}}$, Kang Liao$^{\orcidlink{\orcidauthorB}}$, Chunyu Lin$^{\orcidlink{\orcidauthorC}}$, Tianyi Wei$^{\orcidlink{\orcidauthorD}}$, and Yao Zhao$^{\orcidlink{\orcidauthorE}}$
\thanks{Yinxi He, Chunyu Lin and Yao Zhao are with the Institute of Information Science, Beijing Jiaotong University, Beijing 100044, China, and also with Visual Intelligence +X International Cooperation Joint Laboratory of MOE, Beijing 100044, China. (e-mail: yinxi.he@bjtu.edu.cn, cylin@bjtu.edu.cn, yzhao@bjtu.edu.cn).}
\thanks{Kang Liao and Tianyi Wei are with the College of Computing and Data Science, Nanyang Technological University, Singapore. (e-mail: kang.liao@ntu.edu.sg, bestwty@mail.ustc.edu.cn).}
}


\markboth{Journal of \LaTeX\ Class Files,~Vol.~14, No.~8, August~2021}%
{Shell \MakeLowercase{\textit{et al.}}: A Sample Article Using IEEEtran.cls for IEEE Journals}

\IEEEpubid{0000--0000/00\$00.00~\copyright~2021 IEEE}

\maketitle

\begin{figure*}[ht]
\centering
\includegraphics[width=1.9\columnwidth]{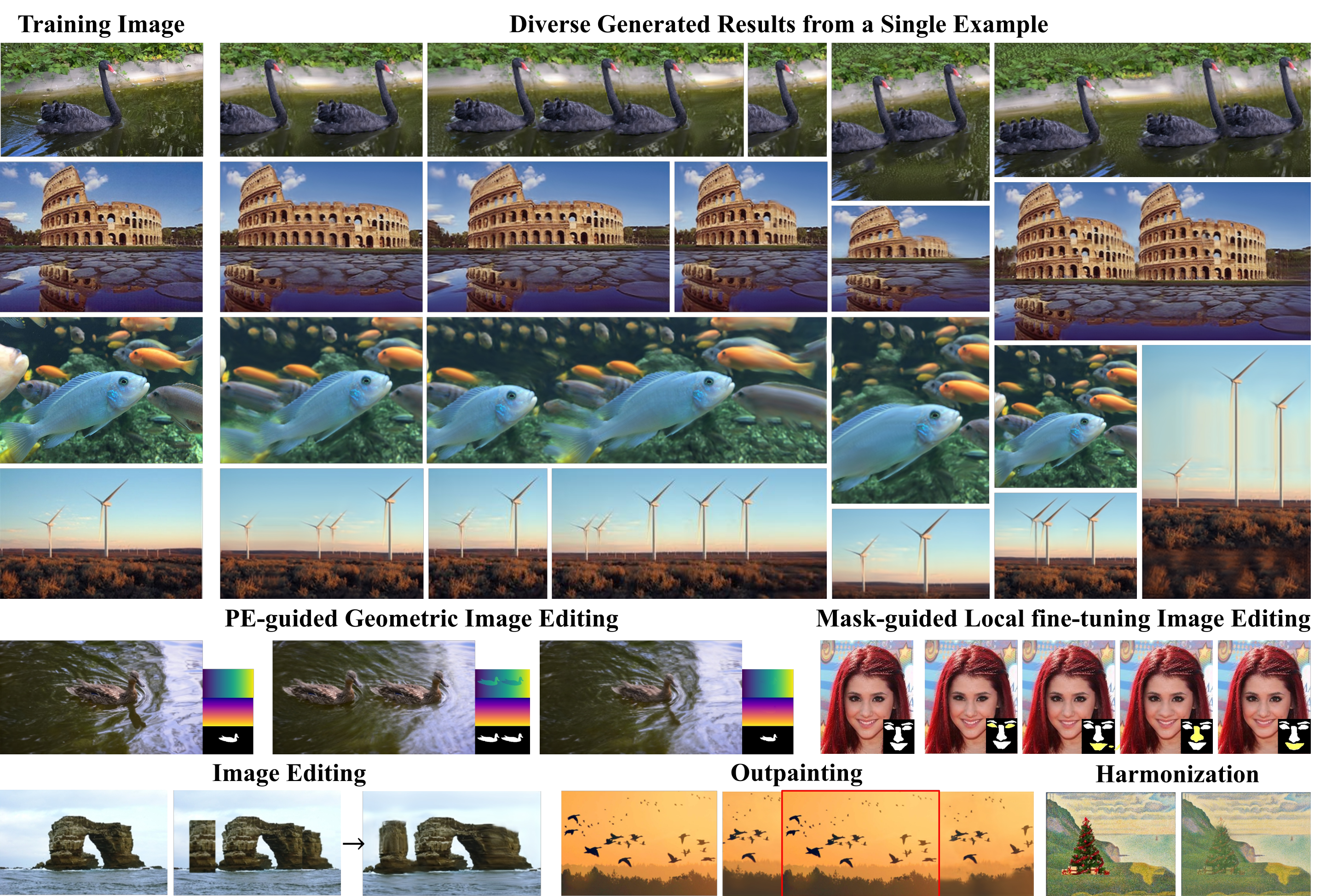}
\caption{StructDiff enables diverse generation and controllable editing from a single training image. \textbf{Top}: Diverse generation at arbitrary scales. \textbf{Middle}: Controllable generation through PE-guided geometric editing and mask-guided fine-tuning. \textbf{Bottom}: Various applications without retraining. Results demonstrate effectiveness in structural preservation and spatial controllability.}

\label{fig:teaser}
\end{figure*}

\begin{abstract}
This paper introduces StructDiff, a generative framework based on a single-scale diffusion model for single-image generation. Single-image generation aims to synthesize diverse samples with similar visual content to the source image by capturing its internal statistics, without relying on external data. However, existing methods often struggle to preserve the structural layout, especially for images with large rigid objects or strict spatial constraints. Moreover, most approaches lack spatial controllability, making it difficult to guide the structure or placement of generated content. To address these challenges, StructDiff introduces an \textit{adaptive receptive field} module to maintain both global and local distributions. Building on this foundation, StructDiff incorporates 3D positional encoding (PE) as a spatial prior, allowing flexible control over positions, scale, and local details of generated objects. To our knowledge, this spatial control capability represents the first exploration of PE-based manipulation in single-image generation. Furthermore, we propose a novel evaluation criterion for single-image generation based on large language models (LLMs). This criterion specifically addresses the limitations of existing objective metrics and the high labor costs associated with user studies. StructDiff also demonstrates broad applicability across downstream tasks, such as text-guided image generation, image editing, outpainting, and paint-to-image synthesis. Extensive experiments demonstrate that StructDiff outperforms existing methods in structural consistency, visual quality, and spatial controllability. The project page is available at \url{https://butter-crab.github.io/StructDiff/}.
\end{abstract}

\begin{IEEEkeywords}
Single-image generation, denoising diffusion probabilistic models, image manipulation.
\end{IEEEkeywords}

\section{Introduction}
\IEEEPARstart{I}{n} recent years, single-image generation has attracted significant attention in the field of generative models. In contrast to the generation methods that rely on large-scale datasets, single-image generation focuses on modeling the internal distribution of one input image. This task trains a model on a single natural image, learning its internal statistics to generate diverse samples with similar visual content, while also supporting various applications such as text-guided generation and image editing.

SinGAN \cite{shaham2019singan} is a pioneering work in this field, which uses multi-scale Patch-GANs to learn hierarchical structure and texture distributions. The follow-up works improved generation quality with diffusion models \cite{kulikov2023sinddm, wang2025sindiffusion}, and expanded to other domains like 3D generation \cite{wu2023sin3dm} and video tasks \cite{nikankin2022sinfusion}. 
However, existing methods face significant challenges in structural preservation. Multi-scale methods such as SinGAN~\cite{shaham2019singan} suffer from error 
accumulation across hierarchical generation. Single-scale methods such as SinDiffusion~\cite{wang2025sindiffusion} struggle with fixed receptive fields that cannot simultaneously capture fine textures and global structures. In addition, spatial controllability remains largely unexplored in single-image generation.

In this paper, we propose StructDiff, a structure-preserving framework based on a single-scale diffusion model. Built on standard formulation of DDPM \cite{ho2020denoising}, StructDiff removes downsampling, upsampling, and attention modules to mitigate the `memorization' issue \cite{nikankin2022sinfusion}, \cite{wang2025sindiffusion} in single-image training caused by overly large receptive fields.
To improve structural preservation, StructDiff introduces an \textit{adaptive receptive field} module, dynamically fusing features from different receptive fields.
This design enables multi-scale structural perception within a single-scale framework. As a result, the model can flexibly adapt to various image types, particularly addressing the common distortion issue in large objects.
To achieve spatial controllability, our StructDiff incorporates a novel 3D positional encoding (PE) as an explicit control signal. Specifically, the positional vector combines spatial coordinates with a binary foreground-background mask, embedding both geometric location and semantic context. The vector is then transformed by a learnable Fourier mapping into high-frequency representations that preserve translation equivariance and sharp semantic boundaries. During inference, by modifying the PE, StructDiff enables spatial control over the generated content, including shifting or scaling specific regions, as well as fine local edits such as facial feature adjustments.

With these designs, StructDiff not only enables diverse random generation at arbitrary scales from a single natural image, but also demonstrates strong applicability in various image generation and editing tasks without retraining. Representative examples are illustrated in Figure~\ref{fig:teaser}. To better assess these capabilities, we further propose a novel evaluation criterion based on large language models (LLMs). This criterion mitigates the inaccuracy of traditional objective metrics in capturing perceptual quality while eliminating the need for heavy labor cost of user studies, offering a new perspective for evaluating single-image generation tasks.

\IEEEpubidadjcol

Recent large-scale diffusion models such as Stable Diffusion~\cite{rombach2022high} and Flux~\cite{labs2025flux1kontextflowmatching} demonstrate strong text-to-image generation capabilities. Through specialized designs, these models can be adapted to single-image tasks. However, training on billions of images makes these models prioritize learned priors over patch statistics of individual images. This limitation leads to introducing irrelevant structures or textures that deviate from the source image characteristics. In contrast, single-image generation methods train exclusively on one image, focusing on capturing its unique internal statistics without interference from external data. This specialization enables superior fidelity in preserving the specific visual patterns of the source image.

Overall, the contributions are summarized as follows. 

\begin{itemize}

    \item We propose StructDiff, a single-image generative framework employing \textit{adaptive receptive field} to enable multi-scale structural awareness in a single-scale diffusion model, significantly enhancing structural preservation. 
  \item We introduce, to the best of our knowledge, the first 3D positional encoding (with Fourier embedding) as an explicit and manipulable spatial prior for single-image generation, enabling precise control over object location, scale, and details.
  \item To bridge the gap between limited objective metrics and labor-intensive user studies, we introduce a novel evaluation paradigm leveraging large language models, offering reliable assessment of generated image quality.
 \end{itemize}

The rest of this paper is organized as follows.
Section~\ref{sec:related-work} discusses related work on  single-image generation and diffusion models.
Section~\ref{sec:method} presents the proposed methodology and the training loss formulation.
Section~\ref{sec:experiment} evaluates our method on three datasets in comparison with several other classic methods and presents the ablation studies.
Finally, Section~\ref{sec:conclusion} concludes the paper.

\begin{figure*}[ht]
\centering
\includegraphics[width=2.0\columnwidth]{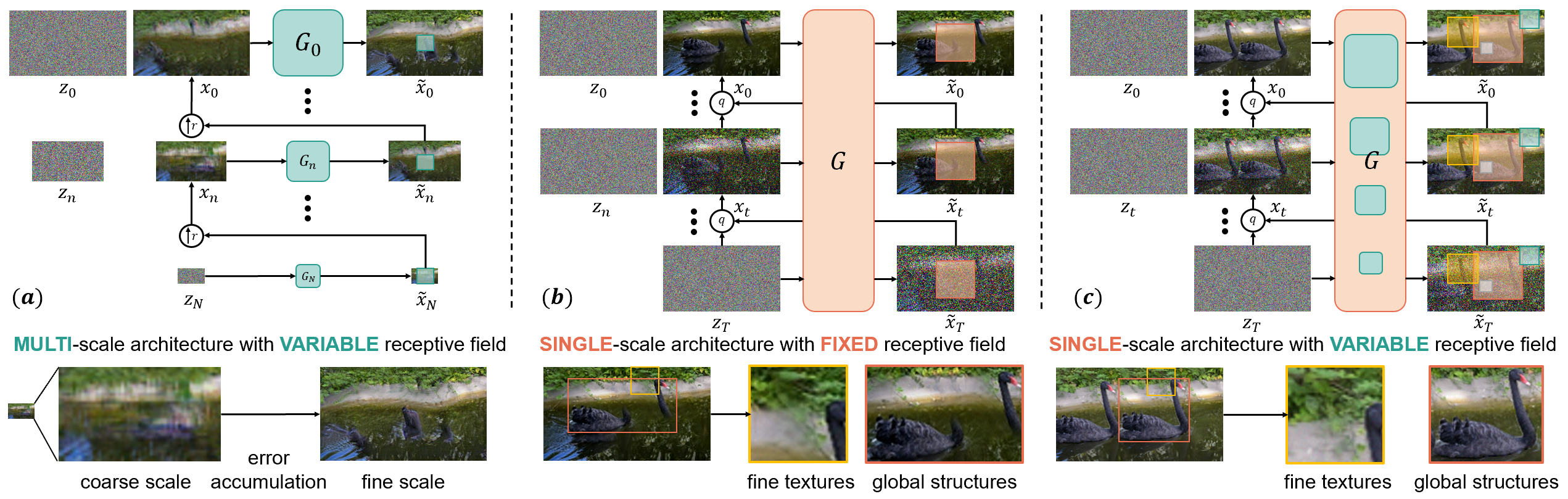} 
\caption{Architectural paradigm comparison in single-image generation. \textbf{(a)} Multi-scale methods train separate generators at different resolutions, enabling multi-scale perception but suffering from error accumulation during hierarchical generation, leading to structural distortions in generated results. \textbf{(b)} Single-scale methods train one generator at fixed resolution, avoiding error accumulation but struggling with fixed receptive fields that cannot simultaneously capture fine textures and global structures, resulting in incomplete structure preservation. \textbf{(c)} Our approach varies receptive field sizes at fixed resolution, achieving multi-scale perception without error accumulation while maintaining structural consistency.}
\label{fig:architectural_comparison}
\end{figure*}

\section{Related Work}
\label{sec:related-work}
\subsection{Single-Image Generation}
Training a model from scratch on a single image is a core direction in internal learning \cite{tirer2023deep}. These methods capture internal distributions to generate diverse samples with similar visual content, without relying on external data. Shaham et al. \cite{shaham2019singan} first proposed an unconditional generation model named SinGAN for single-image tasks using a pyramid-based PatchGAN \cite{isola2017image} architecture. Concurrently, InGAN \cite{shocher2019ingan} introduced a conditional framework focused on image retargeting. Subsequent research explored various improvements for single-image generation. ConSinGAN \cite{hinz2021improved} employed multi-stage parallel training to improve stability and efficiency. PatchGenCN \cite{zheng2021patchwise} proposed a multi-scale energy-based generation framework. GPNN \cite{granot2022drop} learned correspondences between local regions and synthesized new content by cloning neighboring patches. SinDDM \cite{kulikov2023sinddm} replaced PatchGAN \cite{isola2017image} with a more stable denoising diffusion probabilistic model \cite{ho2020denoising}. SinDiffusion \cite{wang2025sindiffusion} further introduced a single-scale diffusion framework tailored for single-image tasks. Other studies extended internal learning \cite{tirer2023deep} to higher-dimensional domains, such as 3D shape generation in Sin3DM \cite{wu2023sin3dm} and video tasks in SinFusion \cite{nikankin2022sinfusion}.

Compared with the above methods, we discard the hierarchical design and instead achieve multi-scale structural perception within a single-scale diffusion framework through an \textit{adaptive receptive} field module.

\subsection{Diffusion Models}
Diffusion models were first presented by Song and Ermon \cite{song2019generative} through score-based generative modeling. Later, Ho et al. \cite{ho2020denoising} proposed denoising diffusion probabilistic models (DDPM), which significantly improved image synthesis quality. Building on this foundation, Dhariwal and Nichol \cite{dhariwal2021diffusion} improved the performance of the model, enabling diffusion models to surpass generative adversarial networks (GANs) \cite{goodfellow2020generative} and autoregressive models \cite{van2016pixel} in image quality. Diffusion models have since emerged as a leading paradigm for high-fidelity image generation. The success of diffusion models has led to rapid expansion in various image-generation tasks, 
including super-resolution (SR3 \cite{saharia2022image}), image translation (Palette \cite{saharia2022palette}), and cross-modal synthesis (Stable Diffusion \cite{rombach2022high}, DALL-E2 \cite{ramesh2022hierarchical}).
Additional research has explored diffusion models for image editing \cite{cao2023masactrl, mou2023dragondiffusion, zhang2023sine, brooks2023instructpix2pix, zhang2024mmginpainting, cao2023difffashion, jiang2024animediff, wang2025stableidentity}, video editing \cite{qin2025truncate, he2025efficient}, 3D content manipulation \cite{chen2025revealing, zhong2025avatarmakeup}, controllable generation \cite{mou2024t2i, zhao2023uni, zhang2023adding, li2023gligen, ju2023humansd, epstein2023diffusion} and personalized synthesis \cite{hu2022lora, gal2022image, ruiz2023dreambooth, ruiz2024hyperdreambooth, zhang2024ssr}, 
highlighting the versatility and strong generative power of diffusion models. Despite their success, most existing diffusion-based methods rely on large-scale training data and task-specific fine-tuning.

\section{Methodology}
\label{sec:method}

\subsection{Overview}

\begin{figure*}[ht]
\centering
\includegraphics[width=2.1\columnwidth]{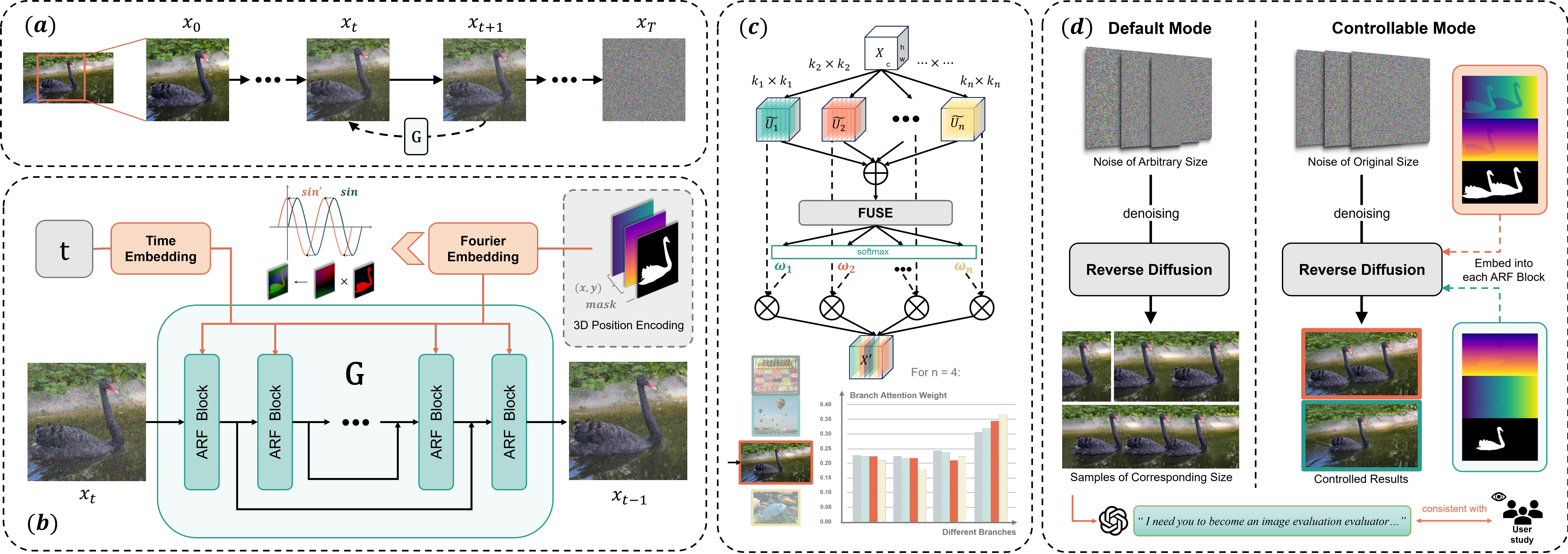} 
\caption{Overview of StructDiff framework. \textbf{(a)} StructDiff trains on randomly cropped patches from a single image to learn internal statistics and generate diverse outputs. \textbf{(b)} The fully convolutional architecture uses Adaptive Receptive Field (ARF) Blocks with residual connections, removing downsampling and attention modules to prevent overfitting. Time embeddings are injected into each block and Fourier-embedded positional encodings enable spatial control. \textbf{(c)} ARF Blocks contain multiple parallel branches with different kernel sizes to extract features at varying receptive fields. The attention mechanism dynamically assigns weights to different branches, with larger receptive fields receiving higher weights when processing images containing large foreground structures, enabling adaptive balance between fine-detail modeling and structure preservation. \textbf{(d)} StructDiff supports default mode for random sampling and controllable mode guided by customized positional encodings, enabling spatial manipulation while maintaining structural coherence.}
\label{fig:pipeline}
\end{figure*}

We introduce StructDiff, a structure-preserving generative framework based on a single-scale diffusion model. An overview of the overall architecture of StructDiff is illustrated in Figure~\ref{fig:pipeline}. It trains on randomly cropped patches from a single natural image \cite{nikankin2022sinfusion} and generates diverse outputs that share similar visual content with the original image. The network architecture follows a modified UNet \cite{ronneberger2015u} design, where downsampling, upsampling, and attention modules are removed to avoid overfitting. StructDiff integrates two core components: an \textit{adaptive receptive field} module for multi-scale structural awareness, detailed in Section~III-B, and a 3D positional encoding for spatial control, detailed in Section~III-C. These two designs enable StructDiff to flexibly handle various types of images, while also supporting interactive spatial control. Notably, StructDiff supports two generation modes. In the default mode, random noise of the desired size produces diverse samples without external guidance. In the controllable mode, customized positional encodings guide the sampling process to produce images with user-specified layouts, as shown in Figure~\ref{fig:pipeline}(d). Overall, StructDiff offers a compact and flexible framework that captures the internal structure of a single image. It also shows strong potential for practical applications.

\subsection{Adaptive Receptive Field}

\noindent \textbf{Motivation: Addressing the Structural Preservation Challenge.} Preserving structural layout remains a fundamental challenge in single-image generation, particularly for images containing large rigid objects. We identify this limitation as rooted in the architectural design choices of existing methods.

Current approaches adopt two distinct architectural paradigms. Multi-scale methods such as SinGAN~\cite{shaham2019singan} and SinDDM~\cite{kulikov2023sinddm} train separate generators at different resolutions, as shown in Figure~\ref{fig:architectural_comparison}(a). Each generator captures features at a specific scale through its effective receptive field. This hierarchical design enables multi-scale perception but introduces error accumulation during the coarse-to-fine generation process. These accumulated errors compromise structural consistency, especially when generating large objects that require precise shape preservation across multiple scales.

Single-scale methods such as SinDiffusion~\cite{wang2025sindiffusion} and SinFusion~\cite{nikankin2022sinfusion} avoid error accumulation by training only one generator at fixed resolution, as shown in Figure~\ref{fig:architectural_comparison}(b). However, this design introduces a different limitation: the fixed receptive field struggles to simultaneously capture fine textures and global structures. Small receptive fields excel at local details but miss large-scale patterns. Large receptive fields capture global context but blur fine details. This fixed-scale perception limits the model's effectiveness across diverse image types.

We propose an alternative solution that addresses both limitations within a single-scale framework, as shown in Figure~\ref{fig:architectural_comparison}(c). Rather than varying image resolution with fixed receptive fields as in multi-scale architectures, we vary receptive field sizes at fixed resolution. This design maintains the computational efficiency and structural stability of single-scale architectures while achieving adaptive multi-scale perception. The key insight is that multi-scale structural awareness does not require multi-scale training. Instead, a single generator can dynamically adjust its perceptual scope based on local image characteristics.

To realize this concept, we introduce an Adaptive Receptive Field (ARF) module. ARF enables the network to adaptively balance fine-detail modeling and large-scale structure preservation based on image content, without introducing the complexity or error accumulation of multi-scale training. The following section details the 
architectural design and technical implementation of this module.

\noindent \textbf{Architectural Design.} 
StructDiff builds on the DDPM~\cite{ho2020denoising} framework and modifies the UNet~\cite{ronneberger2015u} architecture to better support single-image training scenarios. Standard UNet employs downsampling, upsampling, and attention modules that introduce large receptive fields. However, such global feature aggregation tends to cause overfitting in single-image settings by encouraging memorization~\cite{nikankin2022sinfusion}. Recent work~\cite{wang2025sindiffusion} demonstrates that components with global receptive fields correlate strongly with memorization behavior during single-image training. Therefore, StructDiff removes these components and adopts a fully convolutional architecture by stacking multiple ARF Blocks, as shown in Figure~\ref{fig:pipeline}(b). The network maintains a symmetric layout with skip connections between the first and second halves. Sinusoidal time embeddings are injected into each block to ensure consistency across timesteps.

To realize the adaptive multi-scale perception concept introduced above, we design the ARF Block as illustrated in Figure~\ref{fig:pipeline}(c). Each block contains multiple parallel convolutional branches with different kernel sizes following the selective kernel design~\cite{li2019selective}. These branches extract features under varying receptive fields. Smaller kernels focus on local textures while larger ones capture global structures.
The outputs from different branches are then fused and aligned in dimension, followed by a softmax operation that computes attention weights for each branch. These weights are then used to combine the outputs through weighted summation, generating structure-adaptive features. 
As shown in Figure~\ref{fig:pipeline}(c), the visualization of attention weights reveals an adaptive pattern. For images containing large foreground objects, branches with larger receptive fields receive higher weights. This dynamic weight allocation enables StructDiff to adaptively balance fine-detail modeling and large-scale structure preservation.

\subsection{Controllable Generation Driven by PE}

We introduce a positional encoding framework with spatial prior design and Fourier embedding mechanisms.

\noindent\textbf{Spatial Prior Design.} Modeling spatially sensitive structures remains a persistent challenge for current single-image generation methods. This limitation becomes particularly pronounced in tasks requiring strict spatial coherence—such as human face generation—where the synthesized outputs frequently exhibit structural inconsistencies. We train our model on randomly cropped image patches. However, when dealing with large-scale objects, a single patch often fails to capture the complete structural context. This limitation hinders the model's ability to reconstruct the whole object. Furthermore, enabling fine-grained control over localized regions requires the model to understand both aspects: absolute spatial location and semantic context (\textit{e.g.}, foreground vs. background) of each pixel. To address these challenges, we introduce a three-dimensional spatial prior embedding that explicitly encodes both geometric position and semantic foreground-background information. We design the positional embedding (PE) to satisfy two key objectives: (i) to inject location-awareness into each patch, enabling the model to infer its relative position within the global image context; and (ii) to distinguish foreground from background regions, thereby guiding the model in understanding semantic boundaries. Concretely, a 3D positional vector is constructed at each pixel, consisting of spatial coordinates $(x,y)$ and a binary foreground-background mask $ m $.

\noindent\textbf{Fourier Embedding Mechanism.} To support PE-driven spatial control, the embedded representation needs to exhibit two critical properties: translation equivariance and sensitivity to high-frequency details. Translation equivariance helps preserve the consistency of uncontrolled regions and ensures accurate spatial transformation of controlled regions. High-frequency sensitivity is essential for generating sharp boundaries and continuous textures. To meet these criteria, StructDiff applies a learnable Fourier embedding to transform the positional vector:
\begin{equation}e_{fo}\left(x,y,m\right)=\sin\left[B_{fo}\left(x^{\prime},y^{\prime}, m\right)^T\right],\end{equation}
where $x^{\prime}=\textstyle\frac{2x}{W-1}-1 $ and $y^{\prime}=\textstyle\frac{2y}{H-1}-1$ are pixel coordinates, uniformly mapped to the range $[-1, 1]$, and ${B}_{fo} \in  {\mathbb{R}}^{3 \times  n}$ is a learnable weight matrix. On the one hand, the periodic nature of the sine function introduces non-local behavior, allowing the spatial mapping to vary smoothly under translations. On the other hand, the derivative of the sine function is itself a phase-shifted sine wave, preserving the features of the original signal \cite{sitzmann2020implicit}. This enables the model to capture high-frequency spatial information, achieving more precise separation of controlled and uncontrolled regions.

The positional embeddings are injected into every ARF Block in the network. During inference, modifying the positional embedding enables direct control over spatial attributes, allowing operations such as translation, scaling, and localized deformation. These capabilities make StructDiff particularly effective for interactive and fine-grained editing.

\subsection{Training-Free Applications}

StructDiff supports various applications through specialized guidance mechanisms and sampling strategies. The framework adapts to different tasks without requiring retraining or fine-tuning. 

For text-guided generation, StructDiff employs CLIP-based guidance. At selected diffusion steps, a pretrained CLIP model \cite{radford2021learning} evaluates the similarity between the current generated image and the given text prompt. The gradient of the CLIP loss with respect to the predicted reconstruction $\hat{x}_0^{(t)}$ updates the image estimate at each designated time step $t$. The update follows:

\begin{equation}
\begin{aligned}
\hat{x}_0^{(t)} \leftarrow & \eta \, \delta \, m \odot \nabla_{\hat{x}_0^{(t)}} L_{\mathrm{CLIP}} \\
& + (1 - m) \odot (\lambda\hat{x}_0^{(t)}+(1-\lambda)\hat{x}_0^{(t+1)}),
\end{aligned}
\label{eq:text-guidance}
\end{equation}

where $L_{\mathrm{CLIP}}$ measures the semantic discrepancy between the generated image and the text prompt. The term $\nabla{\hat{x}_0^{(t)}}L_{\mathrm{CLIP}}$ represents the gradient, and $m$ is a binary mask specifying the guided regions. The normalization factor $\delta=\|\hat{x}_0^{(t)} \odot m\|_2/\|\nabla_{\hat{x}_0^{(t)}} L_{\mathrm{CLIP}} \odot m\|_2$ controls gradient magnitude, and $\eta \in [0, 1]$ adjusts guidance strength. Notably, the momentum mechanism addresses the issue where each denoiser step tends to undo the preceding CLIP step due to the strong prior of the denoiser. Here, $\hat{x}_0^{(t+1)}$ represents the reconstruction from the previous timestep, and $\lambda \in [0, 1]$ is a momentum parameter that maintains consistency across timesteps. Overall, the gradient guides the sampling trajectory, encouraging alignment with the semantic meaning of the prompt.

For tasks involving reference images, StructDiff uses a sampling strategy inspired by prior work \cite{choi2021ilvr}. A low-pass filter extracts low-frequency structures from the reference image. These structures guide the sampling process by modifying the current state $\hat{x}_t$ at each time step:
\begin{equation}
\hat{x}_t \leftarrow \hat{x}_t - \phi_N(\hat{x}_t) + \phi_N(q(y, t)),
\label{eq:reference}
\end{equation}
where $y$ is the reference image and $q(y, t)$ generates the noisy version of $y$ at timestep $t$. The operator $\phi_N(\cdot)$ represents low-pass filtering. The subtraction removes low-frequency content from the current state, while the addition injects structure from the reference image. These operations align the generation with the reference's global structure.

For outpainting tasks, StructDiff treats content extension as a special case of generation with reference images. The reference image $y$ corresponds to the source image $x_0$ in this context. The method performs seamless extension by injecting the source image into predefined masked regions. A binary mask $m_{\text{in}}$ indicates the inner region where the source image content should be preserved. During specified time steps, the current state $\hat{x}_t$ is updated by directly injecting a noisy version of the source image into the masked region. The noisy version $q(y, t)$ maintains temporal consistency with the diffusion process. The update follows:

\begin{equation}
\hat{x}_t \leftarrow m_{\text{in}} \odot q(y, t) + (1 - m_{\text{in}}) \odot \hat{x}_t.
\label{eq:outpainting}
\end{equation}

This operation ensures consistency in noise levels across the entire image. The injection mechanism allows the outer region to follow the semantics and style of the inner content, resulting in seamless extension of image structures without visible artifacts at boundary regions.

\subsection{Training Loss}

The training objective of StructDiff consists of two components. The first is the standard mean squared error (MSE) loss used in DDPM to denoise the image:
\begin{equation}
\mathcal{L}_{\text{MSE}} = \mathbb{E}_{x_t, x_0, t}\left[\|x_0 - \tilde{x}_{0,\theta}(x_t, t)\|_2^2\right],
\end{equation}
where $x_t$ represents the noisy crop at timestep $t$, $x_0$ is the clean crop of the original image, and $\tilde{x}_{0,\theta}(x_t, t)$ is the predicted crop generated by the model. To improve structural preservation in the foreground region, we introduce a foreground-aware perceptual loss:
\begin{equation}
\mathcal{L}_{\text{FG}} = \mathcal{L}_{\text{VGG}} + \mathcal{L}_{\text{Sobel}},
\end{equation}
where the components of the foreground-aware loss are defined as:
\begin{equation}
\mathcal{L}_{\text{VGG}} = \lambda_1 \cdot \|\phi(\tilde{x}_{0}) \odot m - \phi(x_0) \odot m\|_1,
\end{equation}
\begin{equation}
\mathcal{L}_{\text{Sobel}} = \lambda_2 \cdot \|\text{Sobel}(\tilde{x}_{0}) \odot m - \text{Sobel}(x_0) \odot m\|_1,
\end{equation}
where $\lambda_1,\lambda_2$ control the relative importance of the VGG \cite{simonyan2014very} perceptual loss and Sobel edge loss. $\tilde{x}_{0}$ refers to the predicted crop, $x_0$ is the ground truth crop, and $m$ is the binary foreground mask. The term $\phi$ denotes the VGG-based perceptual feature extractor, and $\text{Sobel}(\cdot)$ represents the gradient edge map. The foreground-aware loss emphasizes perceptual consistency and edge fidelity within the masked foreground region. The total training objective is:
\begin{equation}
\mathcal{L}_{\text{total}} = \mathcal{L}_{\text{MSE}} + \lambda_{FG}\mathcal{L}_{\text{FG}},
\end{equation}
where $\lambda_{FG}$ is empirically set to $1$.

\section{Experiment}
\label{sec:experiment}

\subsection{Implementation Details}
\noindent\textbf{Datasets:} We conduct experiments on three datasets. To evaluate performance across different image types, two custom datasets are employed: Mulmini-N, which contains 20 general images, and Mulmini-L, which consists of 20 images featuring large-scale objects. In addition, Places50, a classic benchmark for single-image generation, is used as a supplementary evaluation.

\noindent\textbf{Evaluation metrics:} To comprehensively evaluate generation performance, this study uses several quantitative metrics. SIFID \cite{shaham2019singan} measures the ability of the model to capture the internal statistics and local feature distributions of the source image. MUSIQ \cite{ke2021musiq} and DB-CNN \cite{zhang2018blind} assess the visual quality of the generated images. LPIPS \cite{zhang2018unreasonable} measures the perceptual diversity between generated samples. In addition, a newly proposed LLM-based scoring method evaluates both image quality (SIQS-G) and structural consistency (SIQS-A). For controlled generation results, PSNR and SSIM \cite{wang2004image} are used to calculate pixel-level differences. PSNR-BG and SSIM-BG evaluate background preservation, while PSNR-FG and SSIM-FG measure foreground structural consistency. These metrics together provide a thorough assessment of both the quality and diversity of the generated images, as well as the model’s ability to maintain structural fidelity and controllability.

\noindent\textbf{Network Configuration:} The experiments are conducted using a 4-branch ARF Block with different kernel sizes: $5 \times 5$, $7 \times 7$, $9 \times 9$, and $11 \times 11$. These configurations capture features at different scales, enabling the model to understand and generate detailed structures.

\begin{figure*}[!htb]
\centering
\includegraphics[width=1.9\columnwidth]{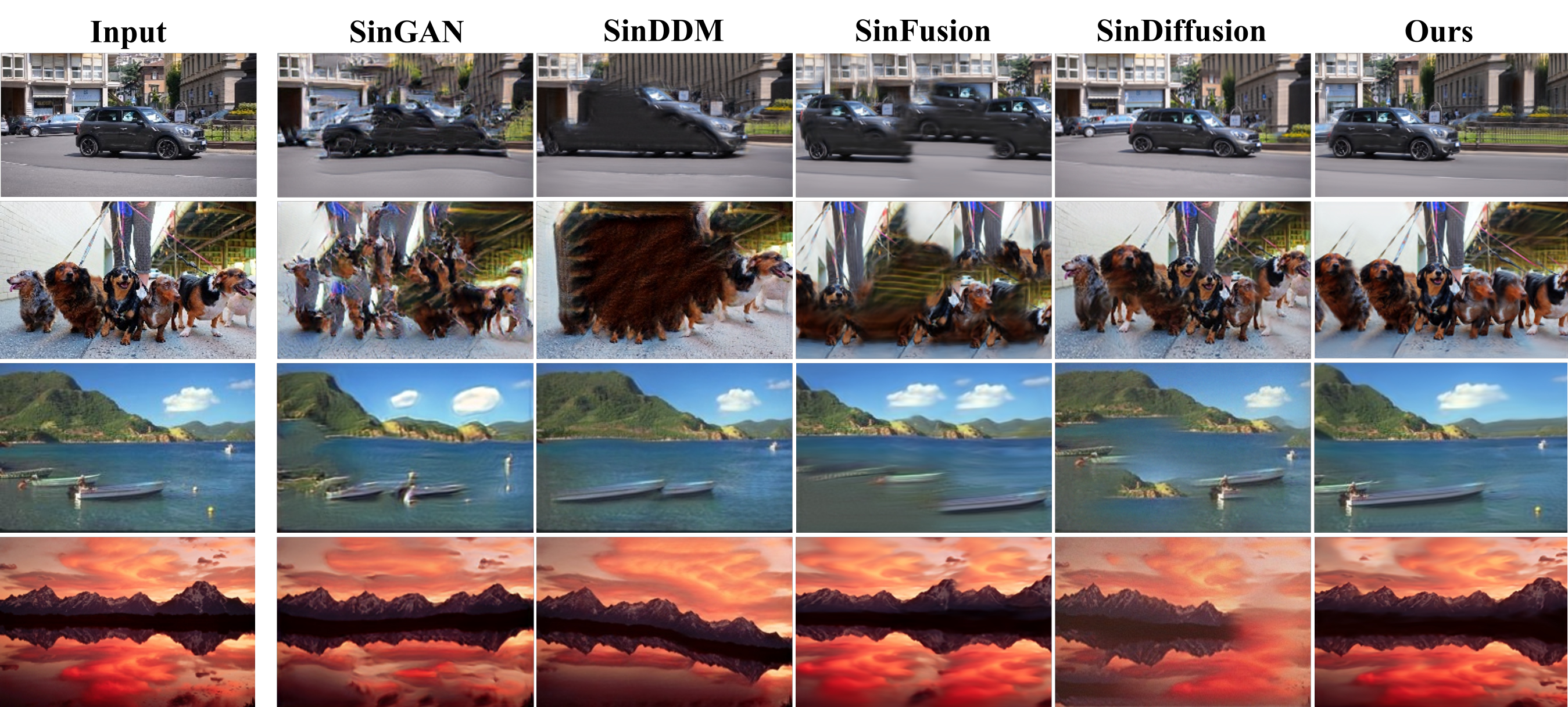} 
\caption{Qualitative comparison with other methods. StructDiff vs. other methods on large-object images (rows 1st-2nd) and natural images (rows 3rd-4th). StructDiff achieves better visual quality and structural consistency across various image types.}
\label{fig:comparison}
\end{figure*}

\begin{table*}[!htb]
\caption{Quantitative comparison on three datasets
}
\centering
\begin{tabular}{ccccccccc}
	\toprule
	\multirow{3.5}{*}{\textbf{Dataset}}&\multirow{3.5}{*}{\shortstack{\textbf{Method}} }&\multicolumn{4}{c}{\textbf{Metrics}}&\multicolumn{3}{c}{\textbf{Point of Scoring (↑)}}\cr
	\cmidrule(l){3-6} 
        \cmidrule(l){7-9}
	&&\multirow{2.1}{*}{SIFID (↓)}&\multirow{2.1}{*}{MUSIQ (↑)}&\multirow{2.1}{*}{DB-CNN (↑)}&\multirow{2.1}{*}{LPIPS (↑)}
	&Generation&Absence of&\multirow{2.1}{*}{Total (5)}\cr
        &&&&&&Quality (2)&Distortion (3)\cr
	\midrule
    
	\multirow{5}{*}{Places50}
        &SinGAN \cite{shaham2019singan}&0.09&46.80&\textbf{0.57}&0.266&-&-&-\cr
	&SinDDM \cite{kulikov2023sinddm}&0.34&44.27&\textbf{0.57}&0.210&-&-&-\cr
	&SinFusion \cite{nikankin2022sinfusion}&0.64&48.40&\textbf{0.57}&0.368&-&-&-\cr
	&SinDiffusion \cite{wang2025sindiffusion}&0.06&46.79&0.56&\textbf{0.387}&-&-&-\cr
	&\textbf{StructDiff(Ours)}&\textbf{0.04}&\textbf{49.01}&\textbf{0.57}&0.311&-&-&-\cr\midrule

    \multirow{5}{*}{Mulmini-N}
    &SinGAN \cite{shaham2019singan}&0.04&45.71&0.57&0.314&1.00&1.60&2.60\cr 
	&SinDDM \cite{kulikov2023sinddm}&0.15&45.09&0.58&0.341&1.40&2.10&3.50\cr 
	&SinFusion \cite{nikankin2022sinfusion}&0.05&48.70&0.59&0.372&1.05&1.45&2.50\cr 
	&SinDiffusion \cite{wang2025sindiffusion}&0.18&46.45&0.57&\textbf{0.403}&1.15&1.85&3.00\cr 
	&\textbf{StructDiff(Ours)}&\textbf{0.03}&\textbf{49.41}&\textbf{0.61}&0.336&\textbf{1.45}&\textbf{2.20}&\textbf{3.65}\cr\midrule
	
	\multirow{5}{*}{Mulmini-L}
    &SinGAN \cite{shaham2019singan}&\textbf{0.06}&48.34&0.63&0.350&0.35&0.65&1.00\cr 
	&SinDDM \cite{kulikov2023sinddm}&0.23&48.87&0.63&0.374&0.90&1.30&2.20\cr 
	&SinFusion \cite{nikankin2022sinfusion}&0.20&52.73&0.61&\textbf{0.429}&0.80&1.15&1.95\cr 
	&SinDiffusion \cite{wang2025sindiffusion}&0.10&51.97&0.61&0.402&1.05&1.55&2.60\cr 
	&\textbf{StructDiff(Ours)}&\textbf{0.06}&\textbf{57.88}&\textbf{0.69}&0.402&\textbf{1.45}&\textbf{2.15}&\textbf{3.60}\cr

	\bottomrule
\end{tabular}
\label{tab:table_total}
\end{table*}

\subsection{LLM-Based Quality Assessment}

Traditional no-reference image quality metrics such as MUSIQ~\cite{ke2021musiq} and DB-CNN~\cite{zhang2018blind} show limited correlation with human perception in single-image generation tasks. Meanwhile, user studies provide reliable results but require substantial time and labor investment. To address these limitations, we introduce a novel evaluation paradigm leveraging large language 
models to achieve reliable quality assessment with reduced cost.

Recent advances in multi-modal large language models have demonstrated strong capabilities in visual understanding and quality assessment. We adopt GPT-4o~\cite{GTP-4o} as the evaluation backbone and design customized prompts tailored to single-image generation characteristics. Unlike general image generation evaluation that focuses on text-image alignment or instruction following, single-image generation requires assessing whether generated content faithfully preserves the structural patterns of the source image while maintaining visual quality.

Our prompt engineering follows a structured approach~\cite{xie2024modification} to ensure evaluation consistency and interpretability. To avoid ambiguous MLLM evaluations, the prompt design must explicitly define evaluation roles, specify analysis dimensions, and quantify scoring criteria. To this end, we position GPT-4o as an ``image generation quality expert'' in the prompt, requiring it to assess generated images from two dimensions: generation quality (2 points), focusing on blur and artifacts, and absence of distortion (3 points), measuring structural consistency. For each dimension, we design an $n$-level rating scale where each score from 1 to $n$ corresponds to a clear quality description. For output format, we adopt Chain-of-Thought prompting, requiring GPT-4o to first describe the observed image features, then provide the reasoning for the score, and finally output the quantitative score. This structured output format enhances evaluation stability and allows human verification of the reasoning process.

\subsection{Unconditional Generation}

Figure~\ref{fig:comparison} presents a qualitative comparison between StructDiff and other methods \cite{shaham2019singan, kulikov2023sinddm, nikankin2022sinfusion, wang2025sindiffusion}. The detailed comparison includes images containing large foreground objects and natural scenes. The results demonstrate that StructDiff consistently outperforms competing methods in overall image quality across various types of images, with particularly pronounced advantages on images containing large foreground objects. StructDiff achieves reasonable reconstructions in both global shape and fine texture details, demonstrating flexible adaptivity and significantly improved multi-scale structural preservation.

In quantitative experiments, we calculated SIFID, MUSIQ, DB-CNN, and LPIPS on Places50 as well as the custom datasets Mulmini-N and Mulmini-L. As reported in Table~\ref{tab:table_total}, StructDiff achieves the best SIFID scores on all datasets, indicating superior ability to capture internal image statistics. For image quality, StructDiff outperforms competing methods on both MUSIQ and DB-CNN, with substantial improvements on Mulmini-L. Although StructDiff does not significantly surpass other methods in LPIPS, we observe a trade-off between diversity and distortion. Figure~\ref{fig:comparison} shows that additional diversity in SinDiffusion \cite{wang2025sindiffusion} often comes with noticeable distortions, deviating from desirable perceptual quality.

For the LLM-based evaluation, as reported in Table~\ref{tab:table_total}, StructDiff achieves the highest scores in both quality and structural consistency. The advantage is especially clear on images with large objects in Mulmini-L, where the average structural consistency score exceeds that of SinDiffusion \cite{wang2025sindiffusion} by 0.6 points out of 3.

\begin{figure*}[ht]
\centering
\includegraphics[width=1.5\columnwidth]{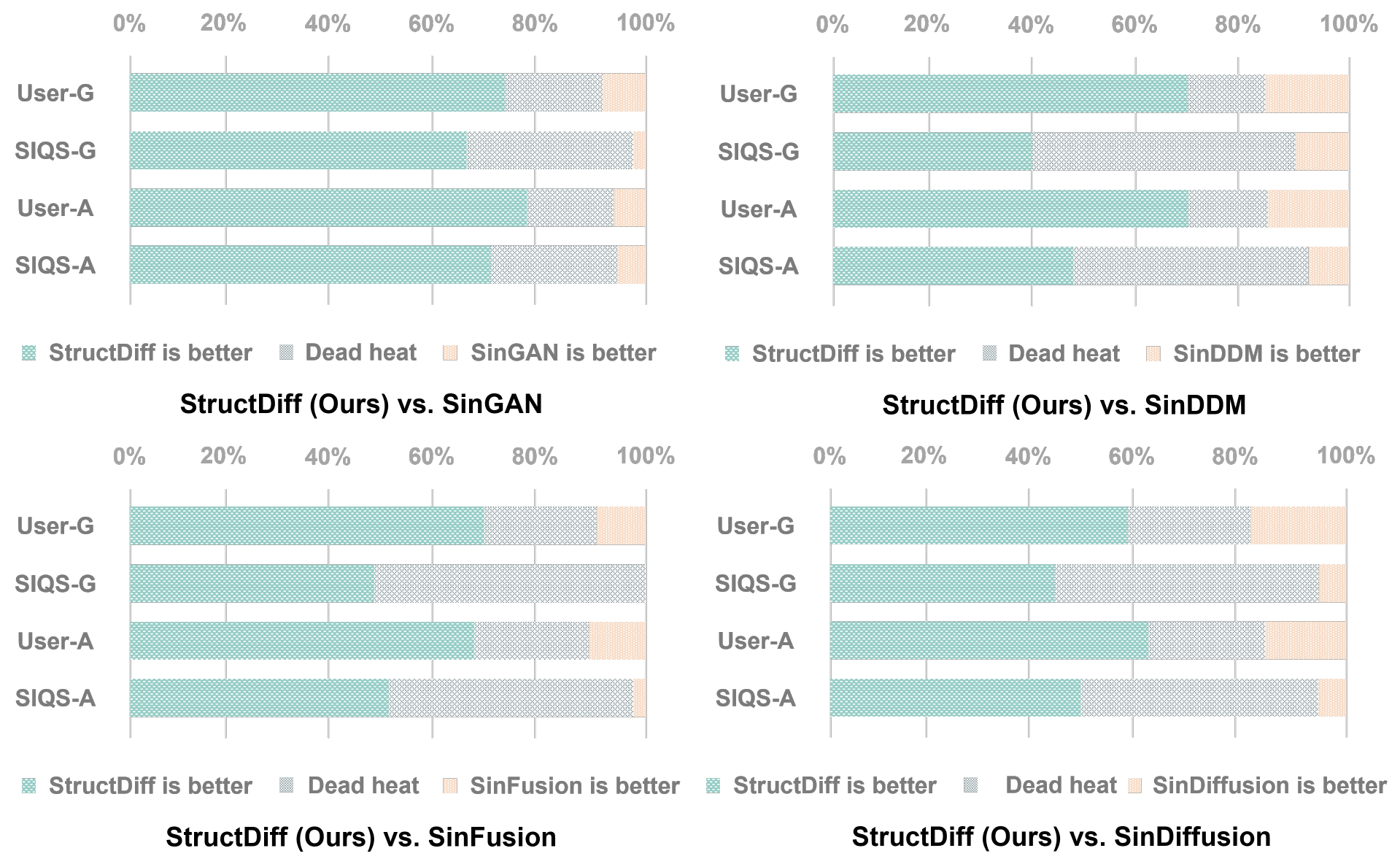}
\caption{Validation of LLM-based evaluation through user study comparison. Overall preference percentages show high correlation between LLM-based scores (SIQS-G, SIQS-A) and user study results (User-G, User-A). Each bar shows the proportion where StructDiff is preferred (\textcolor[rgb]{0.4,0.7,0.6}{teal}), results are tied (\textcolor[rgb]{0.8,0.8,0.8}{gray}), or competing methods are preferred (\textcolor[rgb]{0.9,0.6,0.4}{orange}). The high correlation validates LLM-based evaluation as an efficient alternative to labor-intensive user studies.}
\label{fig:similar}
\end{figure*}

\begin{table}[ht]
\caption{Agreement Analysis Between Metrics and User Study
}
\centering
\begin{tabular}{cccccc}
	\toprule
    &Metric&SIQS-G&MUSIQ&DB-CNN\cr
    \midrule
	&Agreement Rate ($\uparrow$)&\textbf{0.925}&0.706&0.714\cr
	&Cohen's Kappa ($\uparrow$)&\textbf{0.783}&-0.063&-0.057\cr
	\bottomrule
\end{tabular}
\label{tab:table_similar}
\end{table}

To further validate the reliability of LLM-based evaluation, a user study was conducted with 28 volunteers. Each volunteer was randomly shown 20 pairs of images, for a total of 160 pairs, and asked to select preferences based on quality and distortion.
Figure~\ref{fig:similar} compares the LLM-based evaluation results with the user study, showing high correlation at the macro level through overall preference percentages. To evaluate accuracy at the micro level, we introduce the agreement rate and Cohen's Kappa. The LLM-based metric is compared with MUSIQ and DB-CNN, two no-reference image quality assessment methods, on Mulmini-L and Mulmini-N. Results in Table~\ref{tab:table_similar} show that the proposed metric achieves 92.5\% agreement with user study judgments on a per-sample basis, while other metrics show only about 70\% similarity. In addition, a kappa value of 0.789 demonstrates a high level of reliability for the preference consistency.

\subsection{Spatially Controllable Generation}
\begin{figure*}[ht]
\centering
\includegraphics[width=2.0\columnwidth]{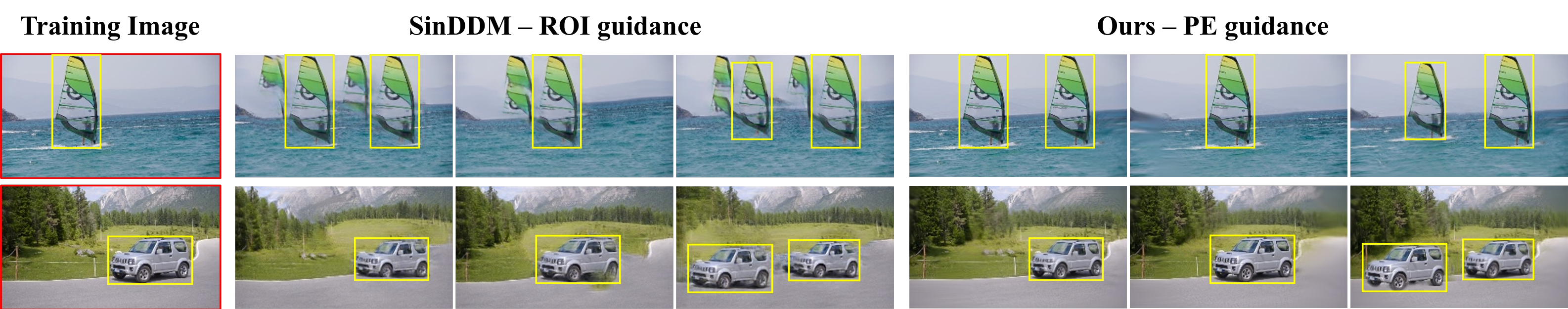}
\caption{Qualitative comparison of spatially controllable generation. StructDiff with positional encoding guidance vs. SinDDM with ROI guidance for controlling foreground objects (\textcolor{yellow}{yellow} boxes). While SinDDM generates blurry edges and artifacts, StructDiff produces sharper edges and maintains background layout. Results validate the effectiveness of 3D positional encoding for spatial control.}
\label{fig:con1}
\end{figure*}

StructDiff also supports spatially controllable generation driven by our designed positional encoding (PE). By manually modifying the positional encoding, users can guide specific regions, enabling transformations such as position shifts, scale changes, and local deformations. Figure~\ref{fig:con1} provides a qualitative comparison between the PE guidance in StructDiff and the ROI guidance in SinDDM \cite{kulikov2023sinddm}.  When controlling the number, scale, and position of foreground objects (marked by yellow boxes), SinDDM generates blurry foreground edges and noticeable artifacts, with poor background consistency compared to the training image. In contrast, StructDiff generates sharper foreground edges, preserves the overall structure, and largely maintains the original layout of the background.

To quantitatively evaluate the preservation of background and foreground structures, PSNR is calculated separately for the background and foreground between the source image and the guided generated image. Table~\ref{tab:table_con} reports the results. StructDiff achieves higher PSNR and SSIM values for both background and foreground compared to SinDDM and other baselines. These results demonstrate that StructDiff achieves better preservation of both background and foreground structures during spatially controllable generation.

\begin{figure}[ht]
\centering
\includegraphics[width=0.48\textwidth]{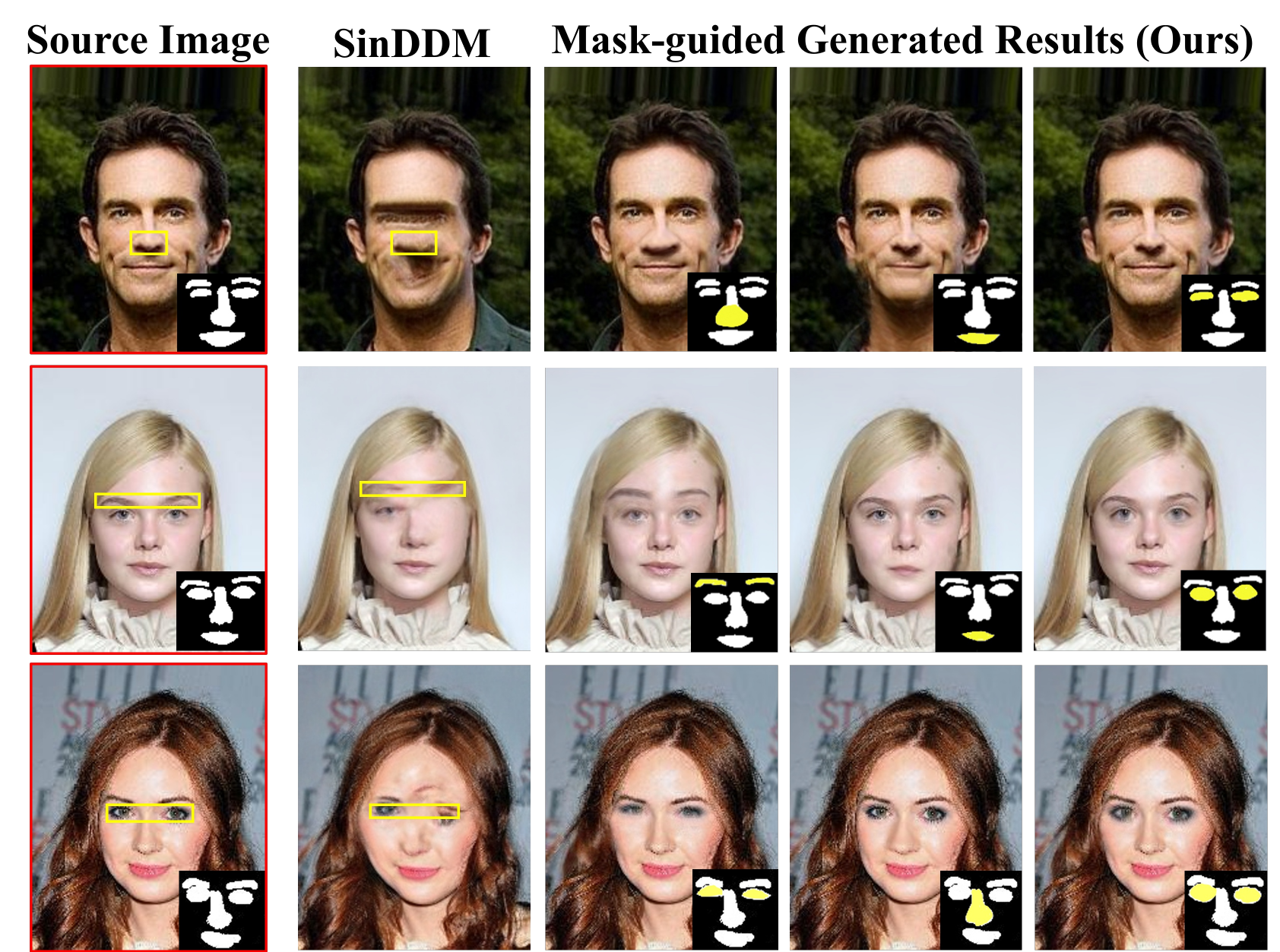}
\caption{Fine-grained local control through mask modification. StructDiff enables precise facial feature adjustments by modifying only the mask component of positional encoding. While SinDDM produces severe distortions, StructDiff achieves intuitive local control while preserving facial structure.}
\label{fig:con3}
\end{figure}

For fine-grained local control, StructDiff does not require modifications to the entire positional encoding (PE). Instead, users can simply modify the mask (the third dimension of PE) to generate the desired image. Figure~\ref{fig:con3} shows that with mask guidance, users can control specific facial features, such as generating wider nostrils or narrower eyes. Meanwhile, SinDDM \cite{kulikov2023sinddm} fails to maintain reasonable facial layouts, producing severe distortions. The results of this local control are visually intuitive and impressive, further demonstrating the flexibility and precision of StructDiff in spatial control.

\begin{figure*}[ht]
\centering
\includegraphics[width=0.97\linewidth]{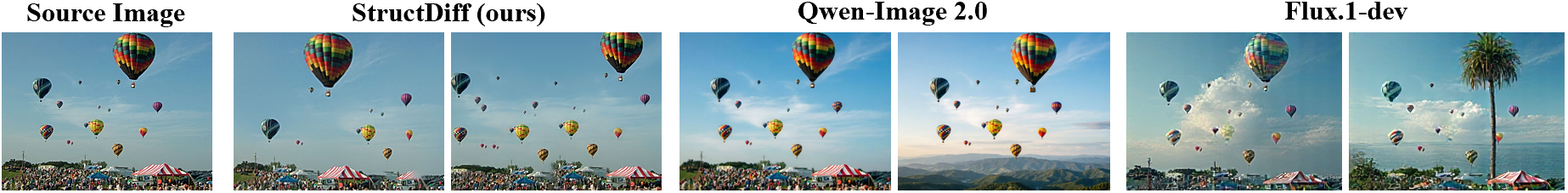}
\caption{Qualitative comparison between DiT-based large models and StructDiff on diverse generation from a single image. While DiT-based models struggle to capture fine-grained internal statistics and exhibit limited diversity or distorted artifacts, StructDiff faithfully preserves local structural patterns and generates highly diverse and visually realistic variations.}
\label{fig:dit_comparison}
\end{figure*}

\begin{table}[ht]
\caption{Quantitative Evaluation of Spatial Control
}
\centering
\begin{tabular}{cccccccc}
	\toprule
    &\multirow{2.7}{*}{Region}&\multirow{2.7}{*}{Metric}&\multicolumn{2}{c}{Method}\cr
    \cmidrule(l){4-5} 
    &&&StructDiff&SinDDM\cr
    \midrule
	&\multirow{2}{*}{Background}&PSNR-BG ($\uparrow$)&\textbf{21.90}&16.62\cr
	&&SSIM-BG ($\uparrow$)&\textbf{0.65}&0.53\cr
    \midrule
    &\multirow{2}{*}{Foreground}&PSNR-FG ($\uparrow$)&\textbf{34.44}&17.94\cr
    &&SSIM-FG ($\uparrow$)&\textbf{0.96}&0.89\cr
	\bottomrule
\end{tabular}
\label{tab:table_con}
\end{table}

\begin{figure*}[ht]
\centering
\includegraphics[width=0.93\textwidth]{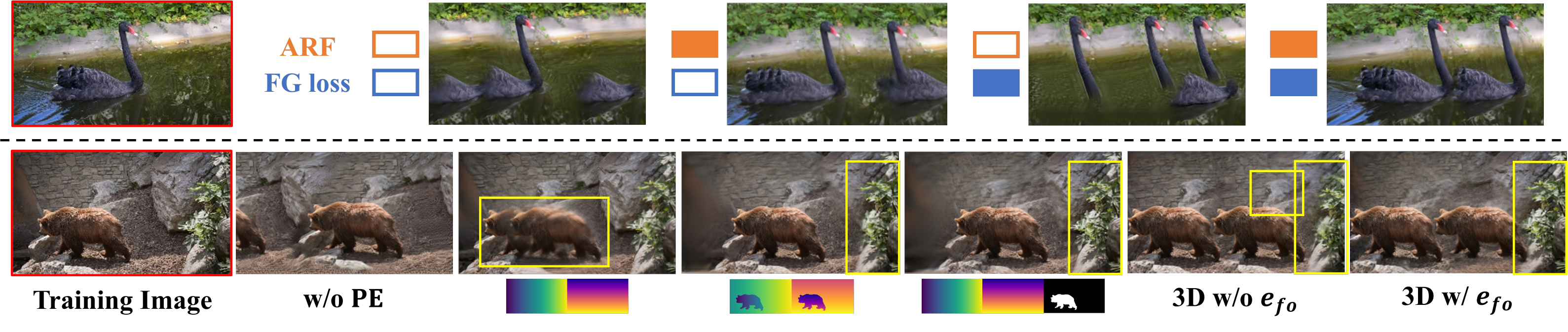}
\caption{Ablation study of StructDiff components. \textbf{Top:} ARF and foreground-aware loss effects. Without both, structures fragment. Combined components achieve coherent structure. \textbf{Bottom left:} Positional encoding strategies. 2D PE causes artifacts when shifting objects. 3D PE with mask enables precise foreground-background separation. \textbf{Bottom right:} Fourier embedding decouples spatial priors, preventing background shifts during foreground translation.}
\label{fig:ablation}
\end{figure*}

\begin{figure*}[ht]
\centering
\includegraphics[width=0.95\textwidth]{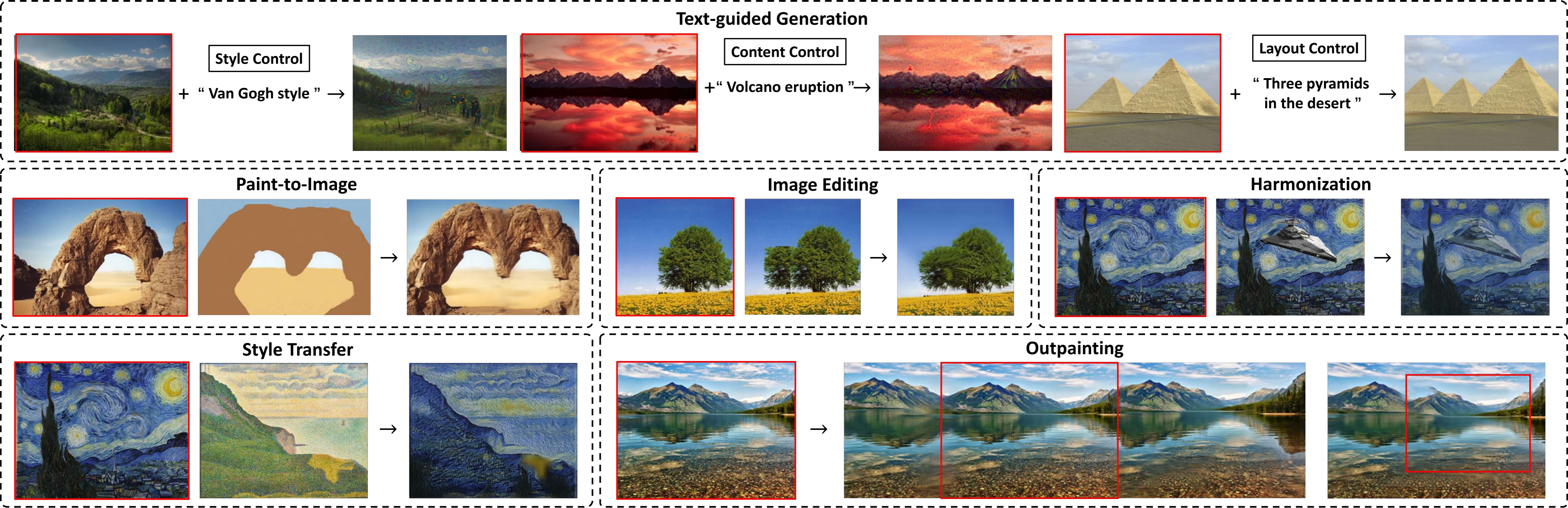}
\caption{StructDiff applications across diverse tasks without retraining. \textbf{Text-guided Generation:} CLIP guidance enables style control, content adaptation, and layout adjustment. \textbf{Paint-to-Image/Editing:} Structure-preserving sampling maintains global structure while allowing local modifications. \textbf{Style Transfer/Harmonization:} Preserves structural coherence during transformation. \textbf{Outpainting:} Seamless content extension maintains visual consistency. Results demonstrate high quality and versatility across image synthesis tasks.}
\label{fig:application}
\end{figure*}

\subsection{Comparison with DiT-based Large Models}

Recent generation models increasingly adopt the Diffusion Transformer (DiT) architecture~\cite{labs2025flux1kontextflowmatching, wu2025qwenimagetechnicalreport}. These models achieve outstanding performance on massive datasets through global self-attention mechanisms. However, this advantage becomes an inherent limitation in internal learning~\cite{tirer2023deep}, especially for single-image generation scenarios. To be specific, the self-attention mechanism in DiT captures global dependencies across the entire image, which is precisely the ``memorization'' problem~\cite{nikankin2022sinfusion, wang2025sindiffusion} we aim to avoid in single-image generation.

Figure~\ref{fig:dit_comparison} presents a qualitative comparison between StructDiff and DiT-based large models on diverse generation from single images. The results demonstrate that current DiT-based models exhibit limitations in capturing fine-grained image structure information. They tend to either output rigid copies of the original image or generate poor results with limited diversity. In contrast, StructDiff explicitly models internal patch statistics, successfully avoiding global memorization tendencies. This design preserves local structures, making it highly suitable for single-image generation tasks.

\subsection{Ablation Study}

To evaluate the effectiveness of key components in StructDiff, we conduct ablative experiments on three aspects: the \textit{adaptive receptive field module (ARF)}, the foreground-aware loss ($\mathcal{L}_{\text{FG}}$), and various positional encoding strategies for controllable generation.

The first analysis examines the impact of ARF and the foreground-aware loss on structural consistency. Figure~\ref{fig:ablation} top row shows that removing both components results in fragmented structures. The swan’s neck and body appear as separate regions, disrupting continuity. ARF alone enables broader context perception but weak spatial relationships. Foreground-aware loss alone increases foreground fidelity but minimally improves structure. Combining both components achieves coherent global structure.

The next analysis compares positional encoding (PE) strategies. Figure~\ref{fig:ablation} bottom left shows that the model generates diverse but uncontrolled samples without PE. Standard 2D coordinate encoding enables position-aware generation. However, shifting the foreground, such as moving the bear 30 pixels to the right, produces blurred artifacts at the original location due to missing positional values. Using two separate coordinate systems for foreground and background allows object relocation but causes background compression and deformation, as indicated by the yellow box. This result suggests interference between the translated foreground and the static background. The 3D positional encoding uses a mask to explicitly distinguish foreground and background regions, enabling more precise separation and improving the consistency of spatial guidance. This design also supports fine-grained local editing using masks, as shown in Figure~\ref{fig:con3}.

The final analysis investigates the effect of Fourier embedding. Figure~\ref{fig:ablation} bottom right shows that foreground translation causes background shifts without Fourier embedding, indicating entangled spatial priors. Fourier embedding decouples foreground and background, enabling more precise control.

The ablation study demonstrates the importance of each proposed component. The ARF module and foreground-aware loss significantly improve the preservation of object structure. The 3D positional encoding with Fourier embedding enables flexible and accurate spatial manipulation.

\subsection{Applications}

StructDiff demonstrates strong flexibility across various image generation and editing tasks. The method achieves high-quality results in text-guided generation, image editing, paint-to-image, style transfer, harmonization, and outpainting without requiring retraining or fine-tuning.

Text-guided generation uses CLIP-based guidance as described in~(\ref{eq:text-guidance}). The gradient guidance enables style manipulation, content adaptation, and layout adjustment, aligning generated content with text prompts. Tasks involving reference images benefit from the structure-preserving sampling strategy outlined in~(\ref{eq:reference}). Paint-to-image, image editing, harmonization, and style transfer all leverage this approach. 
Results show strong alignment between generated content and reference image structure. Outpainting achieves seamless content extension through the specialized sampling process defined in~(\ref{eq:outpainting}). The method maintains visual consistency and follows the original image semantics across extended regions, generating coherent results without visible artifacts at transition areas.

Figure~\ref{fig:application} illustrates representative examples from these applications. The results confirm high visual quality across diverse generation scenarios. StructDiff proves effective not only for single-image generation but also serves as a general framework for image synthesis across practical tasks.

\begin{figure}[ht]
\centering
\includegraphics[width=0.42\textwidth]{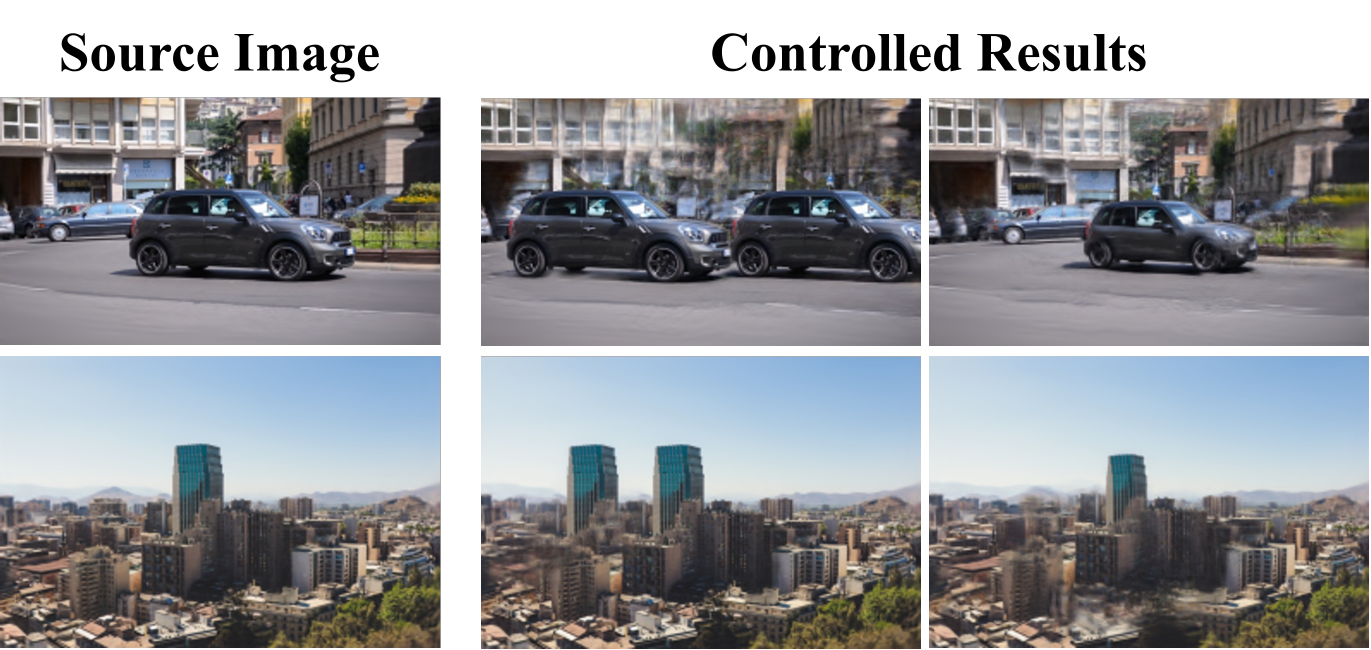}
\caption{Failure cases of StructDiff. The method struggles with foreground-background separation in complex backgrounds and cannot complete occluded parts as humans would expect due to limited semantic information.}
\label{fig:limitation}
\end{figure}

\subsection{Limitation Discussion}
We show some failure cases in Figure~\ref{fig:limitation}, indicating that StructDiff faces challenges in certain scenarios. When performing controlled generation on images with complex backgrounds, the method sometimes confuses foreground-background separation, leading to chaotic background textures in the generated results. Additionally, when moving occluded foreground objects from behind occluding elements, the method cannot complete the hidden parts as humans would expect. This limitation occurs because the model is trained on a single image and lacks semantic information. The model can only rationalize the occluded parts based on the distribution learned from the source image, as shown in the figure where the building occlusion is maintained rather than revealing the complete object.
These limitations suggest potential directions for future improvement. Curriculum learning represents a promising approach to address these challenges. This strategy mimics human learning processes by introducing a training progression from simple to complex scenarios. This progressive training paradigm could enhance scene understanding without overwhelming the model, potentially improving performance in challenging scenarios while maintaining the advantages of single-image generation.

\section{Conclusion}
\label{sec:conclusion}

We propose StructDiff, a single-image diffusion model that tackles the dual challenges of structural preservation and spatial control. By integrating \textit{adaptive receptive field} modules and 3D positional encoding priors, StructDiff generates high-quality, diverse outputs while maintaining structural layout, especially for images with large objects. Notably, the positional encoding approach represents the first exploration of PE-based spatial manipulation in single-image generation. To complement existing evaluation methods, we further introduce a novel LLM-based protocol for assessing generation quality and controllability. Extensive experiments demonstrate StructDiff's superior performance over existing methods across a variety of image generation and editing tasks.

\bibliographystyle{IEEEtran}
\bibliography{references}

\newpage


 




\end{document}